# Why KDAC? A general activation function for knowledge discovery

Zhenhua Wang[1], Haozhe Liu[1], Fanglin Liu[2], Dong Gao[1*]

[1]Beijing University of Chemical Technology; [2]Southwest Jiaotong University

{2019210490, 2019210489, gaodong}@mail.buct.edu.cn; LF0721@my.swjtu.edu.cn

*Abstract*: Deep learning oriented named entity recognition (DNER) has gradually become the paradigm of knowledge discovery, which greatly promotes domain intelligence. However, the current activation function of DNER fails to treat gradient vanishing, no negative output or non-differentiable existence, which may impede knowledge exploration caused by the omission and incomplete representation of latent semantics. To break through the dilemma, we present a novel activation function termed KDAC. Detailly, KDAC is an aggregation function with multiple conversion modes. The backbone of the activation region is the interaction between exponent and linearity, and the both ends extend through adaptive linear divergence, which surmounts the obstacle of gradient vanishing and no negative output. Crucially, the non-differentiable points are alerted and eliminated by an approximate smoothing algorithm. KDAC has a series of brilliant properties, including nonlinear, stable near-linear transformation and derivative, as well as dynamic style, etc. We perform experiments based on BERT-BiLSTM-CNN-CRF model on six benchmark datasets containing different domain knowledge, such as *Weibo*, *Clinical*, *E-commerce*, *Resume*, *HAZOP* and *People's daily*. The evaluation results show that KDAC is advanced and effective, and can provide more generalized activation to stimulate the performance of DNER. We hope that KDAC can be exploited as a promising activation function to devote itself to the construction of knowledge.

## 1. INTRODUCTION

Activation function provides generalization for deep learning and can serve different tasks with superb presentation. For example, ReLTanh can improve the processing of fault diagnosis for planetary gearboxes and rolling bearings [1], GEV is leveraged to develop a COVID-19 diagnostic model [2], xUnit is dedicated to dealing with image restoration problems [3], and PSAF is suitable for wind power forecasting [4], etc. There is no doubt that activation functions have great research potential for application scenarios.

In the era of big data, deep learning oriented named entity recognition (DNER) with great generalization and scalability has overwhelmingly replaced the previous methods, and achieved knowledge discovery, which is of irreplaceable significance to promote intelligent development such as decision-making support and product services, etc. [5]. DNER draws support from the activation function to learn the complex latent representation, which can freely extract knowledge from text. The excavated knowledge has promising, meaningful and broad application value, such as text summary [6], life event detection [7], content filtering and monitoring [8], trend analysis [9], surface webs and social networks mining [10, 11], biological and patent knowledge retrieval [12, 13], false news identification [14], crime pattern analysis [15], criminal network development [16], criminal organization member detection [17], network security survey [18], adverse drug reactions [19], rumor interception [20], medical diagnosis [21], biomedical literature management [22], geological exploration [23], user preference insights [24, 25] and soft skill community construction [26]. Where, the research [21] can help doctors understand the patient's condition concisely and provide favorable conditions for diagnosis and treatment timely. The research [23] based on Mexican News has developed the first Mexican Geoparser. The researches [24, 25] can explore users' opinions, preferences or content insights which are hidden in an unstructured way, and build an online recommendation system by enhancing personalization with the help of Twitter. The research [26] can detect shared job descriptions and build soft skill communities through automatic text mining.

Nowadays, the activation functions embedded in DNER mainly included in Appendix 1, which is our survey on the DNER research between 2020 and 2021 from some representative journals in *Elsevier*, including *Neurocomputing* (*Neu*), *Neural Networks* (*NN*), *Expert Systems with Applications* (*ESA*), *Information Processing & Management* (*IPM*) and *Knowledge-Based Systems* (*KBS*). According to the investigation, our work is the first to consider the activation function for knowledge discovery. We briefly summarize the related aspects of the DNER research, such as the main models and the activation functions. It can be concluded that Tanh, Sigmoid and ReLU

* Corresponding author
E-mail address: gaodong@mail.buct.edu.cn (D. Gao).

dominate all the DNER work except for one case under SELU (see Fig.1, Fig.2 and Appendix 1).

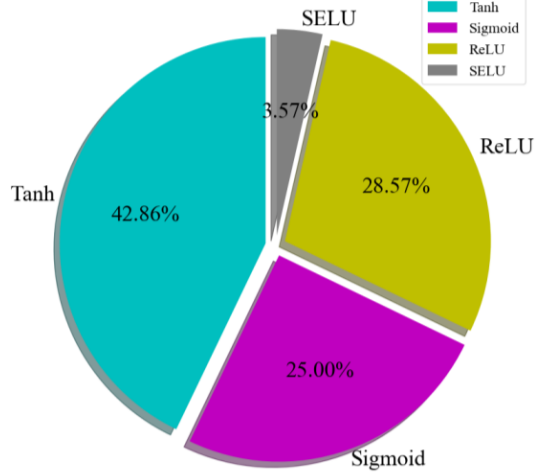

Fig.1: A brief survey of activation functions on DNER researches from representative journals.

(1) Tanh [27], is a smooth activation function with the value range of (-1, 1) that can be differentiable everywhere. In the current research, it is mostly emerged in models with BiLSTM. Nevertheless, the neurons in the activation regions on both sides of Tanh are prone to saturation, which causes the gradient to vanish and impedes the back propagation of the network.

(2) Sigmoid [28], is a smooth activation function with the value range of (0, 1), which is mainly applied to the gated threshold. For example, the LSTM family regards the zero-value activated via Sigmoid as a forgetting mechanism to erase the transmitted information. However, like Tanh, it also has gradient vanishing. Besides, its output value is always greater than zero.

(3) ReLU [29], retains the original input in the positive part and sets that of the rest to zero, it has superiorities in convergence speed, screening effect and alleviating gradient vanishing activated over the positive region. Yet, ReLU itself also is provided with some fatal deficiencies, it forces the zero output to paralyze the negative region (although this operation is conducive to sparse activation), which causes the neurons unable to be awakened in perpetuity and the negative output is eliminated. More unfortunately, ReLU is not differentiable at (0, 0).

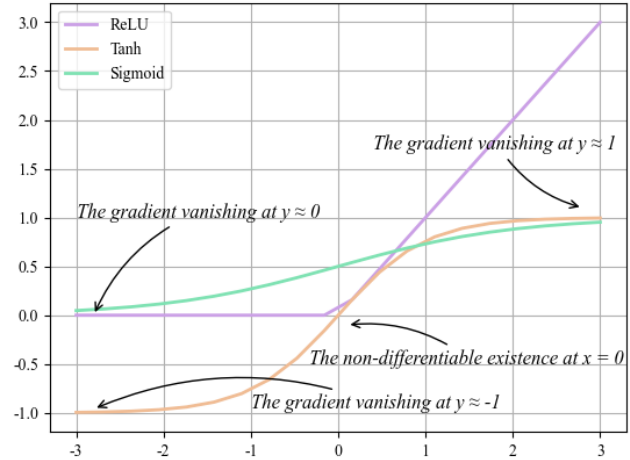

Fig.2: The activation functions of DNER. Where, ReLU is non-differentiable at $x = 0$, the gradient of Tanh vanishes at $y \approx \pm 1$, the gradient of Sigmoid vanishes at $y \approx 1$ and $y \approx 0$.

An intuitive common sense is that knowledge is frequently susceptible, and minor text differences can also implicate the meaning it represents. So, these activation functions preferred by DNER enjoy some fatal deficiencies, which may bring some obstacles to knowledge discovery, mainly:

(1) The incomplete semantic representation, caused by the gradient vanishing. When the gradient disappears, the weight update for hidden layers away from the output will be less ideal, it may even be almost equivalent to the initialization, which poorly encourages the disturbance of parameters. Therefore, the learning of some neurons in DNER is insufficient, which makes the mining of text semantics may not be exquisite enough. As emphasized above, even the slight roughness of a text vector may cause its semantics to change, which makes the model have the divergence in understanding knowledge. Despite the problem is usually ignored by previous studies, we refuse to take it lightly.

(2) The omission of latent semantics, caused by the non-differentiable existence and no negative output. Specifically, the parameter optimization is accompanied by gradient updating, which requires that the activation function, the vinculum of knowledge transmission, to be differentiable everywhere. Significantly, ReLU loses derivability at the zero point, which may reject the data fitting and make some important features omitted. To repair this defect, the sub-gradient [30] is introduced in the actual operation, and the derivative at the piecewise point is set to zero. This operation is only the expedient, since it powerlessly redeems the loss of latent semantic features. Subsequent researches [31] skillfully avoid the irrationality of the zero point, but there are still non-differentiability and other weaknesses at other activation sites. More urgently, for Sigmoid and ReLU, their output lacks negative values, the network may be fragile in the process of training, which discards numerous features and introduces some unnecessary errors [32]. Previous studies [31] have considered the lethality of no negative value, yet there are some limitations.

All in all, DNER has promoted the orderly progress of the current era with knowledge discovery, and how to improve DNER is extremely critical. So, what is in urgent demand is a universal activation function that can mitigate the negatives faced by previous studies.

In this study, a general **ac**tivation function for **k**nowledge **d**iscovery termed KDAC with the superiority of surmounting the above troubles is proposed. To the best of our knowledge, there are no relevant works dedicated to this research, and our work comes to fill in this gap.

Detaily, KDAC draws inspirations from linear Newton interpolation for nonlinear eigenvalue problem, and leverages an approximate smoothing algorithm to manage non-differentiable existence. Besides, KDAC absorbs Tanh form exponential structure and adjustable linear structure to treat gradient vanishing and no negative output. KDAC has a series of gratifying properties. It can activate various modes, change dynamically and freely convert diverse inputs, to respond to different types of knowledge, its nonlinearity ensures the fitting ability, its near-linear transformation and derivative can guarantee the stability of the knowledge mapping. Experiments on six domain datasets indicate that KDAC can provide more efficient activation performance, it can be exploited and given priority for knowledge discovery. The main highlights of this research are as follows.

(1) A general activation function for knowledge discovery termed KDAC is proposed.
(2) Mode transformations and approximate smoothing algorithms are embedded into KDAC.
(3) KDAC conquers the gradient vanishing, the non-differentiable existence and no negative output.
(4) Multiple experiments indicate that KDAC is effective, advanced and generalized.

Section 2 is the related work, which mainly reviews DNER and its main models. Section 3 analyzes the deduction and properties of KDAC. Section 4 presents processes and results of experiments. Discussion and conclusion are expounded in Section 5 and Section 6, respectively.

## 2. RELATED WORK

What we shall be explicit about is what is the specific pattern of knowledge discovery implemented by DNER and its main models, which helps our work be treated in good faith.

### 2.1. DNER

The named entity in DNER refers to the knowledge slice with specific meaning or strong representativeness in the text [33], such as the equipment and the material. Formally, given a text sequence $S = <c_1, c_2, ..., c_n>$, DNER outputs a series of lists $<I_b, I_e, k, t>$, where, each list represents a named entity; $I_b$ and $I_e$ with the definition domain $[1, n]$ are the begin index and the end index of the entity respectively; $k$ is the entity, $t$ is the type of the entity. It is worth noting that how to pursue the correctness of knowledge is the primary concern, because some sophisticated applications can hardly bear the misleading caused by wrong knowledge. An intuitive case is shown in Fig.3, where, the input is '氨酸性气进入非加氢酸性水汽提，可能造成人员中毒，污染环境 (ammonia acid gas enters the non-hydrogenated acid water stripping, which causes personnel poisoning and environmental pollution) ', DNER is committed to exporting as correctly as possible: {Ammonia acid gas: Material; non-hydrogenated acid water stripping: Equipment; personnel poisoning: Consequence; environmental pollution: Consequence}.

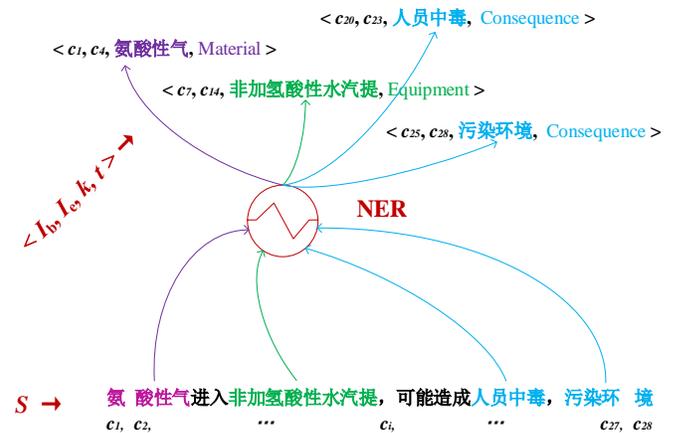

Fig.3: A simple case to illustrate DNER. DNER aims to automatically extract knowledge fragments, that is, it mainly pursues the correctness of $<I_b, I_e, k, t>$.

The method for knowledge discovery has mainly gone through the following stages [33]:

(1) Early methods were mainly based on rule templates. Researchers constructed a series of templates through the structural information of the text, and extracted the expected entities in the way of grammar rules and regularity. For example, "association between parts of speech" is one of the feasible rules. Yet, the rule template is coupled to the current information with specific dictionaries and syntactic patterns, which treats new knowledge entities weakly and has great limitations in predictability. Besides, the meaning and expression of knowledge in different fields are various, this method fails to be migrated and applied to new domains and unknown entities, its generalization and scalability are rather shallow. Further, for large-scale irregular text, the implementation of this method requires cumbersome, complex and huge workload. At present, it is mainly used as an auxiliary.

(2) Subsequently, traditional machine learning was leveraged to explore knowledge. This method usually converts the text into a series of vector representations, such as TF-IDF, and explores knowledge by mining and processing the representation features, such as clustering and Hidden Markov algorithm, which can internally inspire the data and liberate the manual labor. Unfortunately, this method frequently obtains the surface meaning such as word frequency, and fails to capture the semantic information of the text, it is restricted in feature learning. Therefore, the prospect of this method is not expected, and it weakly adapts to the confusion and influx of massive data, let alone some complex or sophisticated fields.

(3) Nowadays, the breakthrough of deep learning theory has contributed to the popularization of artificial intelligence, DNER has overcome the previous limitations and gradually become the paradigm of knowledge discovery. This method generates a nonlinear mapping from input to output through the activation function, which can learn sophisticated features from irregular data. Furthermore, DNER is mainly in the form of end-to-end, which can diminish the feature propagation error and undertake more diverse and rich mining designs. However, how to improve the performance of DNER for enhancing the quality of knowledge faces great challenges. The existing researches are mainly carried out from the perspective of network structure and data preprocessing, but less attention is paid to the activation function. Therefore, the design of activation function in DNER is a valuable and potential research.

Summarily, DNER can deal with the issues existing in previous knowledge discovery. It achieves promising, persistent and generalized expectations in exploring knowledge, becomes a paradigm with a more efficient mechanism, and shines brightly in application services with practical value. Inevitably, how to design an appropriate activation function to promote DNER is a research worth looking forward to.

### 2.2. Model

At present, the models that govern DNER include BiLSTM, CRF and CNN (see Appendix 1 and Fig. 4). In addition, BERT, as one of the NLP paradigms, has the attributes of Transformer and Attention, hence, BERT can also be regarded as the main model. The following is a brief description.

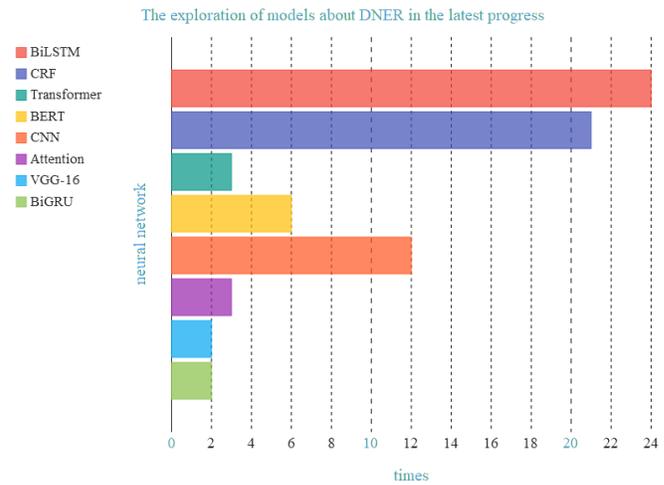

Figure 4: A brief survey of models in DNER researches from the representative journals

**BiLSTM**: Long short-term memory (LSTM) is a particular recurrent neural network that can alleviate the problems of gradient vanishing and gradient explosion. Through the specifically designed gate structure with the ability to remember and forget information, it can selectively save context information and address the dependency issue. BiLSTM is composed of a pair of LSTMs for bidirectionally mining semantics between texts, which can reliably collect long-distance information features, hence, it is very suitable for processing text data and is the default for various NLP practices.

**CRF**: The decoder can predict the probability of different tags to which each entity belongs. Generally, if the tags with the highest probability obtained by *softmax* are directly used as the final prediction, some incorrect tag sequences will be generated. Interestingly, the object processed by CRF is not a single character, but an entire sequence, so the correlation between tags can be weighed to alleviate this problem. CRF can add some constraints automatically learned from the training data in probability calculation to ensure the effectiveness of prediction labels. Therefore, CRF is usually considered as a general decoder for sequence tasks.

**CNN**: The network mainly extracts character-based feature vectors in the coding stage. Briefly, CNN completes local feature mining with character level granularity by means of convolution operation, which can not only enrich features, but also alleviate the impact of word segmentation errors. The convolution window grabs the k-gram fragment information in each input text in turn, different k-gram fragments usually mean different character level features.

**BERT**: It is a pre-training language model with deep bidirectional Transformers based on the Attention [34], and trained from massive text through Next sentence prediction (NSP) and Masked Language Model (MLM) tasks for language understanding. Where, NSP can obtain the representation of sentence level by predicting whether the two sentences are connected, which enables BERT to understand the relationship between sentences. MLM is similar to cloze filling, in short, it randomly covers and replaces the tokens in each training sequence with a probability of 15%, and then predicts the original words at that position. BERT contains rich semantic information and can serve DNER well through fine-tuning, and BERT based neural networks have become the popular solutions of DNER [35-38], which has greatly promoted the vigorous development of NLP.

## 3. The proposed activation function – KDAC

KDAC is devoted to knowledge discovery, it gets rid of the problems existing in previous activation functions, which can perfect the incomplete representation of semantics and the omission of latent semantics. In this section, we elaborate it from three aspects, as follows.

### 3.1. Deduction of KDAC

To be rigorous, we demonstrate the deduction of KDAC with details, rather than directly presenting the specific expression. Please see below.

Although the way that different regions are embedded with different activations is conducive to the performance of the activation function, an unavoidable fact is that this operation inadvertently introduces non-differentiable existence. Take ReLU, the *Max* function, as a case, when $x < 0$, ReLU$(x) = 0$, when $x \geq 0$, ReLU$(x) = x$. Obviously, the first derivative of ReLU$(x)$, ReLU'$(x)$, is available when $x \to 0^-$ or $x \to 0^+$, but it has no value at $x = 0$. Similarly, the common activation function has no derivative at the piecewise point. For this problem, inspired by the research of Chen et al. [39], in some cases, we can draw inspirations from the idea of nonlinear eigenvalue problem, see Equ.1.

$$F(\psi)x = 0 \tag{1}$$

Where, $\psi$ is the approximation of the derivative at the piecewise point, $F(\psi)$ is the analytical function about $\psi$, for example, if Equ.1 is the standard eigenvalue problem, then $F(\psi) = \psi I - A$, $A$ is the assumed derivative at the piecewise point. Note that the form conforming to $F(\psi)$ is unknown, which may be the polynomial eigenvalue problem or the delayed eigenvalue problem, etc.

Considering that for the piecewise point $x = k$ of the general activation function, the derivatives at $x \to k^-$ and $x \to k^+$ are available, the nonlinear function $F(\psi)$ is approximated by linear Newton interpolation between these two known points, see Equ.2.

$$F(\psi) = F(k^-) + (\psi - k^-)F(k^-, k^+) + R(\psi, k^-, k^+) \tag{2}$$

Where, $F(k^-, k^+) = [F(k^-) - F(k^+)] / (k^- - k^+)$, $R(\psi, k^-, k^+)$ is the remainder. Chen et al. [39] have applied the successive linear Newton interpolation method without the remainder to the nonlinear eigenvalue problem, and given the proof and rationality. Therefore, in view of the above, we can get Equ.3 in the functional, that is, the approximate analytical function.

$$P(\xi, f_i, f_j) = f_i + \xi(f_i - f_j) \tag{3}$$

Where, $\xi \in (0, 1)$ is the switching factor, $f_i$ and $f_j$ are two different functions, $P(\xi, f_i, f_j)$ is the piecewise function. Our purpose is to approximately make $P(\xi, f_i, f_j)$ differentiable in a smooth form. When $\xi \to 0$, $P(\xi, f_i, f_j) \to f_i$; when $\xi \to 1$, $P(\xi, f_i, f_j) \to f_j$. Obviously, the abscissa of the intersection between $f_i$ and $f_j$ can be obtained by solving the equation $d_{ij} = f_i - f_j = 0$. We set the interval $-\mu < d_{ij} < \mu$ as the range to be smoothed, and $P(\xi, f_i, f_j)$ under the action of *Max* function at this range can be converted into $P(\mu, f_i, f_j)$, at the same time, $\xi$ can be rewritten as Equ.4, co-regulated by $\mu, f_i$ and $f_j$.

$$\xi = \frac{1}{2} + \frac{f_j - f_i}{2\mu} \tag{4}$$

Equ.5 is obtained by calculating the first derivative of Equ.3.

$$\frac{dP}{dx} = \frac{df_i}{dx} + \frac{d\xi}{dx}(f_j - f_i) + \xi(\frac{df_j}{dx} - \frac{df_i}{dx}) \tag{5}$$

For the boundary conditions, that is, when $\xi = 0$, $f_i - f_j = \mu$, and when $\xi = 1$, $f_i - f_j = -\mu$. Equ.6 and Equ.7 can be obtained respectively.

$$\frac{dP}{dx} = \frac{df_i}{dx} + \mu\frac{d\xi}{dx} \tag{6}$$

$$\frac{dP}{dx} = \frac{df_j}{dx} - \mu\frac{d\xi}{dx} \tag{7}$$

To achieve that $P(\xi, f_i, f_j)$ under the two boundary conditions has equal derivatives, the term $(\mu \cdot d\xi / dx)$ needs to be eliminated. Hence, the compensation term $\tau$, Equ.8, is considered to be added to the right of Equ.5, see Equ.9, and see Equ.10 for the original function of Equ.9.

Now, we have obtained the approximate smoothing form of $P(\xi, f_i, f_j)$ under the load of *Max* function, that is, Equ.10, it can eliminate non-differentiability of piecewise points.

$$\tau = 2\mu\xi\frac{d\xi}{dx} - \mu\frac{d\xi}{dx} \tag{8}$$

$$\frac{dP}{dx} = \frac{df_i}{dx} + \frac{d\xi}{dx}(f_j - f_i) + \xi(\frac{df_j}{dx} - \frac{df_i}{dx}) + \tau \tag{9}$$

$$P_{Max}(f_i, f_j) = f_i + \xi(f_j - f_i) - \mu\xi^2 + \mu\xi \quad (10)$$

Similarly, The $P(\varsigma, f_i, f_j)$ under the load of *Min* function can be approximated to the Equ.11.

$$P_{Min}(f_i, f_j) = f_i + \varsigma(f_j - f_i) + \mu\varsigma^2 - \mu\varsigma$$
$$\varsigma = \frac{1}{2} + \frac{f_i - f_j}{2\mu} \quad (11)$$

Considering that Tanh, Sigmoid and ReLU dominate DNER with their respective strengths, we assign the two functions, Tanh and $y = x$, to the two parameters of $P_{Min}$, since we consider that the choice of Tanh (see Equ.12) is more representative than that of Sigmoid (see Equ.13), see Equ.14. Considering that the gradient of the negative region may vanish under the activation of $P_{Min}$, we nest it into $P_{Max}$ and assign $y = x$ to another parameter, see Equ.15.

$$Tanh(x) = (e^x - e^{-x})/(e^x + e^{-x}) \quad (12)$$

$$Sigmoid(x) = 1/(1 + e^{-x}) \quad (13)$$

$$P_{Min} = P_{Min}(Tanh, x) \quad (14)$$

$$P_{Max} = P_{Max}(P_{Min}(Tanh, x), x) \quad (15)$$

Encouraged by ReLU variants [40], we design trainable factors called $\beta_i$ ($\beta_i > 0$) to play a generalization for Equ15, $\beta_i$ can adjust the amplitude of the two linear functions via the transmitted information, see Equ.16.

$$KDAC(x) = P_{Max}(P_{Min}(Tanh, \beta_1 x), \beta_2 x), \beta_i \in \mathbb{R}^{b \times n \times e} \quad (16)$$

Equ.16 is the mathematical expression of KDAC activation function (see https://github.com/pyy-copyto/KDAC for the code and see Algorithm 1 for the procedure), where $\beta_1 \neq \beta_2$, $b$ is the batch size, $n$ is the number of steps, $e$ is the embedding size.

Now, we have completed the deduction of KDAC, it evolves approximately from linear Newton interpolation for nonlinear eigenvalue problem, and is accompanied by the smoothing elimination of non-differentiable points, the overcoming of gradient vanishing and the enhancement of generalization. Next, we analyze its properties.

---

**Algorithm 1: KDAC**

**Input**: Tensor $X$
**Parameter**: Variable: $\beta_1, \beta_2$. Constant: $\mu$
**Output**: Tensor $Y$
1: **for** $x$ **in** $X$ **do**
2:   $f_1, f_2, f_3 = \beta_1 \cdot x$, Tanh($x$), $\beta_2 \cdot x$
3:   $\varsigma = 0.5 - 0.5 \cdot (f_2 - f_1)/\mu$
4:   **for** $\varsigma \leftarrow 0$ to 1 **do**
5:     $P_{Min} = f_1 \cdot (1 - \varsigma) + \varsigma \cdot f_2 - \mu \cdot \varsigma \cdot (1 - \varsigma)$
6:     $\xi = 0.5 + 0.5 \cdot (f_3 - P_{Min})/\mu$
7:     **for** $\xi \leftarrow 0$ to 1 **do**
8:       $P_{Max} = P_{Min} \cdot (1 - \xi) + \xi \cdot f_3 + \mu \cdot \xi \cdot (1 - \xi)$
9: **return** $Y = P_{Max}$

---

### 3.2. Property of KDAC

As is known to all, each activation function has its own strengths in dealing with different tasks, which is hard to be evaluated unilaterally. In this subsection, we show a series of worthwhile and interesting aspects of KDAC by activation curve, nonlinearity, near-linear transformation, analysis of parameters and derivative analysis, the details are as follows.

#### 3.2.1. Activation curve

Fig.5 is the schematic curve of KDAC, which contains three different groups of $\beta_i$ ($\beta_i > 1$) for intuitive display.

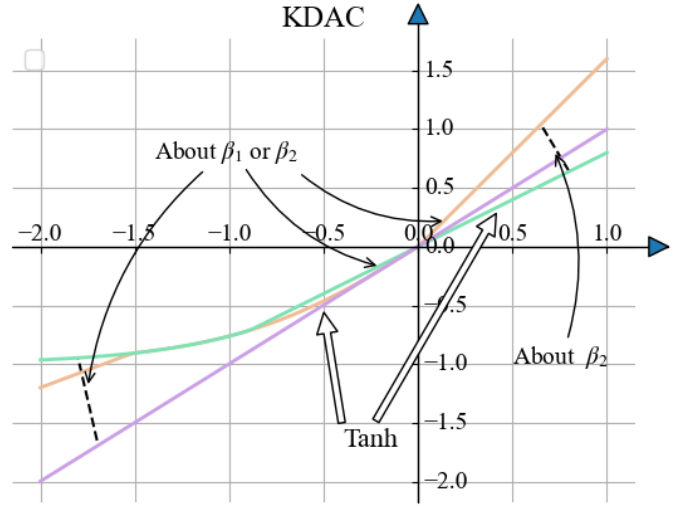

Fig.5: A schematic diagram case of KDAC under multiple switching factors. Where, $\beta_1 > 1$, $\beta_2 > 1$, $\beta_1$ monitors the negative region and $\beta_2$ dominates the positive region, and Tanh is the coupling between them.

It can be distinctly seen that KDAC is mainly composed of three activation segments, which are regulated by $\beta_1$, Tanh and $\beta_2$ successively, obviously, the coupling of the three is dynamic, which can adaptively treat various input. From a macro perspective, KDAC is a smooth and monotonic curve with stable rise, there is no saturations in the whole activation region. Besides, the linear structure on both sides of KDAC ensures the diversity of output values. Further, the design principle of KDAC naturally eliminates non-differentiable points. Thus, KDAC can get rid of the dilemma of gradient vanishing, no negative output and non-differentiable existence, which can alleviate the omission and incomplete representation of potential semantic information caused by output anomalies.

### 3.2.2. Nonlinearity

Nonlinearity is the critical part, which ensures that the neural network will not degenerate into linear operation with low generalization. With the demonstration in Section 3.1, KDAC can be embedded in a variety of forms, and its entire derivative is not a fixed constant. Hence, KDAC satisfies nonlinearity.

### 3.2.3. Near-linear transformation

Near-linear transformation means that the output value will not change significantly with the input difference [41]. In addition, this stable training is beneficial to the performance of the model. We will explain this characteristic of KDAC through step-by-step analysis. It should be considered that to control variables and facilitate analysis and display, we fix the parameter $\mu$, the radius range of non-differentiable points to be smoothed, as a number close to 0, ignore its state, that is, KDAC is almost equivalent to the $Max(Min(Tanh, \beta_1 x), \beta_2 x)$, the details about the $\mu$ are recorded in section 3.2.4.

Firstly, we analyze the $P_{Min}(Tanh, \beta_1 x)$ and let it as $H(x)$, let $G(x) = \beta_1 x – Tanh$. Obviously, when $y = x$, $G(x)$ has only one zero-point $x = 0$, that is, $\beta_1 = 1$. There are two parts below, see Fig.6.

(1) $\beta_1 \geq 1$:

when $x > 0$, $G(x) > 0$, $H(x) = Tanh$; when $x < 0$, $G(x) < 0$, $H(x) = \beta_1 x$.

(2) $0 < \beta_1 < 1$:

$G(x)$ has two additional zero point and set them as $x = k$ and $x = -k$, respectively. Where $k$ can be estimated approximately, and with the decrease of $\beta_1$, $k$ is infinitely close to $(1 / \beta_1)$. So, when $k > x > 0$, $H(x) = \beta_1 x$; when $x > k$, $H(x) = Tanh$; when $0 > x > -k$, $H(x) = Tanh$; when $x < -k$, $H(x) = \beta_1 x$.

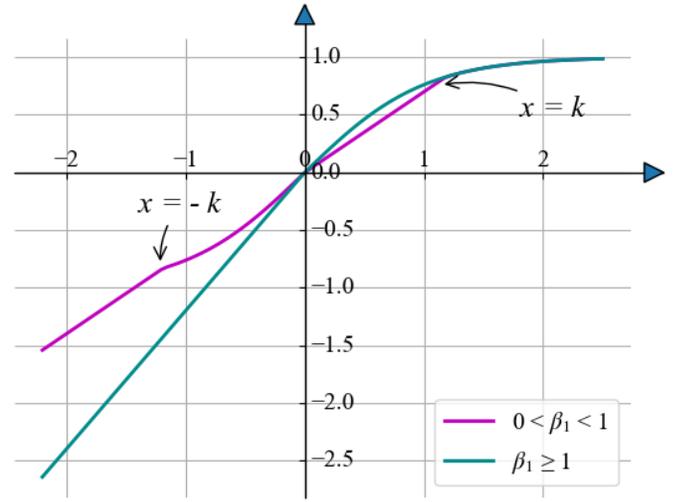

Fig.6: The schematic diagram of $H(x)$.

Secondly, we explore the $P_{Max}(H(x), \beta_2 x)$ and let it as $Q(x)$, let $M(x) = \beta_2 x – H(x)$, Note that $\beta_1 \neq \beta_2$. The following is the discussion from the positive activation region and the negative activation region, see Fig.7.

(1) $x \geq 0$:

Evidently, in the positive activation region, $H(x)$ tends to $y = 1$ regardless of $\beta_1$. Hence, with the increase of input $x$, $M(x) > 0$, $Q(x) = \beta_2 x$, which satisfies the near-linear transformation.

(2) $x < 0$:

With the decrease of input $x$, the negative activation region is the comparison of two linear functions. If $\beta_1 x > \beta_2 x$, $M(x)$ is equal to $\beta_1 x$, otherwise $M(x)$ is equal to $\beta_2 x$. Therefore, this is also consistent with the near-linear transformation.

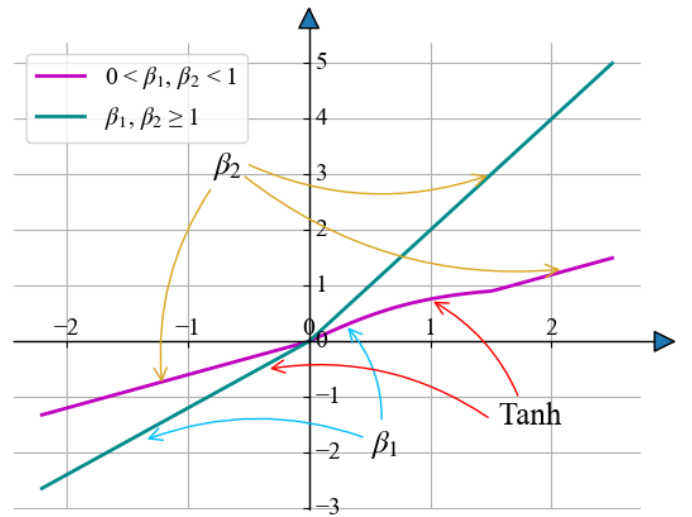

Fig.7: The schematic diagram of $Q(x)$.

To sum up, we prove the near-linear transformation of KDAC.

### 3.2.4. Analysis of parameters

$\beta_i$ and $\mu$ are two parameters of KDAC, which have a direct impact on the trend of KDAC, where, $\beta_i$ ($\beta_1$ and $\beta_2$) regulates the interference region of Tanh, and $\mu$ regulates the approximate smoothing range of non-differentiable points. See below for details.

For parameter $\beta_i$, its analysis has been described in Section 3.2.1 and Section 3.2.3. It can be noticed that the effect of $\beta_1$ or $\beta_2$ in interval (0, 1) or interval [1, +∞) on KDAC is inconsistent, which is reflected in the spread of Tanh and the linear structure in the negative activation region, that is, the exponential structure of Tanh can emerge in the negative activation region and also in the positive activation region (see Fig.5 or Fig.7). Besides, the linear structures cooperating with Tanh are different, for example, KDAC assigns $\beta_1$ to the left of Tanh, $\beta_2$ to the right part, or $\beta_1$ or $\beta_2$ to both sides of Tanh, and so on.

In addition, another interesting thing worth to be discussed is the change of KDAC caused by the value comparison between $\beta_1$ and $\beta_2$, the following are the details.

We continue the analysis in Section 3.2.3, $\mu \approx 0$, $H(x) = P_{Min}(\text{Tanh}, \beta_1 x)$, $M(x) = \beta_2 x - H(x)$, $G(x) = \beta_1 x - \text{Tanh}$, $Q(x) = P_{Max}(H(x), \beta_2 x)$, take $\beta_1 < \beta_2$ as a case.

(1) $0 < \beta_1, \beta_2 < 1$. when $k > x > 0$, $H(x) = \beta_1 x$; when $x > k$, $H(x) = \text{Tanh}$; when $0 > x > -k$, $H(x) = \text{Tanh}$; when $x < -k$, $H(x) = \beta_1 x$, see the purple curve in Fig.6. Then, under the interference of $P_{Max}$, the activation of interval (-∞, -k], (-k, -t] and (-t, +∞) are converted to $\beta_1 x$, Tanh and $\beta_2 x$, respectively, where, $t$ is the absolute value of the zero of $M(x)$ except $t = 0$, it can be estimated approximately, and with the decrease of $\beta_2$, $t$ is infinitely close to $k$, see Fig.8.

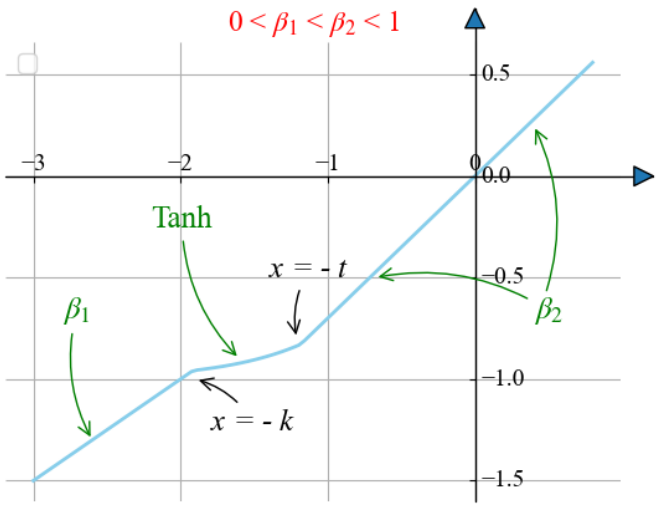

Fig.8: The trend of KDAC when $0 < \beta_1, \beta_2 < 1$.

(2) $0 < \beta_1 < 1, \beta_2 \geq 1$. $M(x)$ has only one zero point, that is, $x = 0$, in which case it cannot interfere with the interval (-∞, 0]. Therefore, the negative region follows $H(x)$ and the positive region follows $\beta_2 x$, see Fig.9.

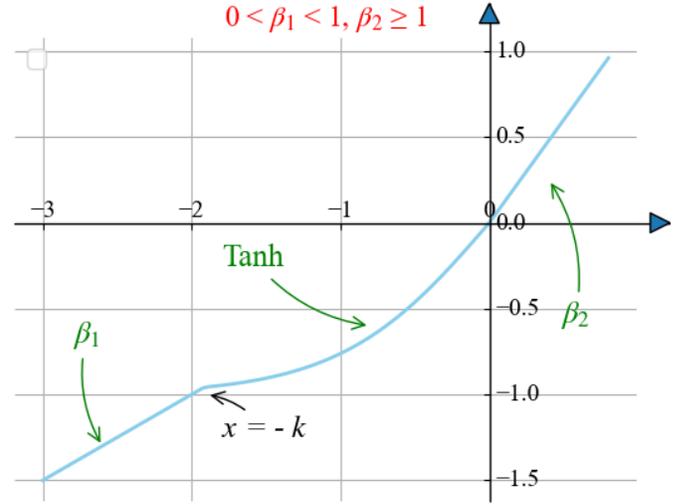

Fig.9: The trend of KDAC when $0 < \beta_1 < 1$ and $\beta_2 \geq 1$.

(3) $\beta_1, \beta_2 \geq 1$. Both $M(x)$ and $G(x)$ have only one zero, that is, $x = 0$. Hence, $H(x)$ let Tanh and $\beta_1$ fill the positive and negative regions respectively, and then $Q(x)$ let $\beta_2$ replace Tanh in the positive region, see Fig.10. It can be perceived that the KDAC in this case is like a ReLU variant with variable parameters, but what's better is that KDAC can eliminate the non-differentiable point with the help of the parameter $\mu$.

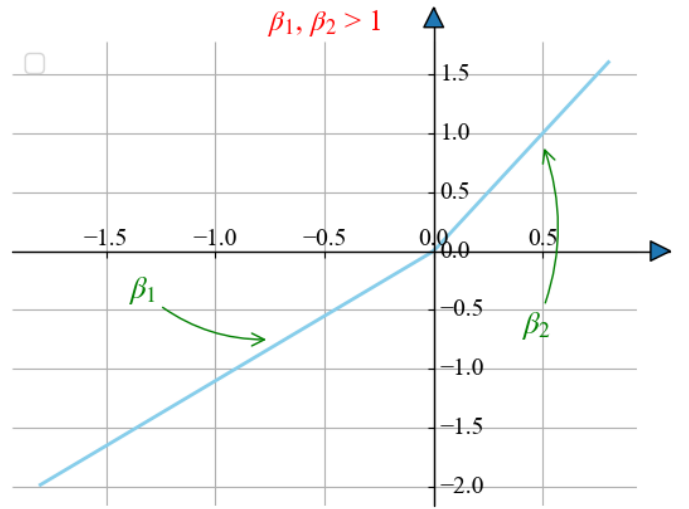

Fig.10: The trend of KDAC when $\beta_1, \beta_2 \geq 1$.

Similarly, in the case of $\beta_1 > \beta_2$, KDAC also has the above multiple changes, and one of the intuitive changes is that the interference of Tanh transfers from the negative region to the positive region, see Fig.11.

All in all, KDAC can change dynamically with the help of parameter $\beta_i$, which is reflected in 1): the interaction forms of linear and Tanh are diverse, and 2): the exponential structure can emerge in different regions. So, KDAC can ensure generalization, and freely convert diverse inputs, that is, it has the capability to respond to different types of knowledge.

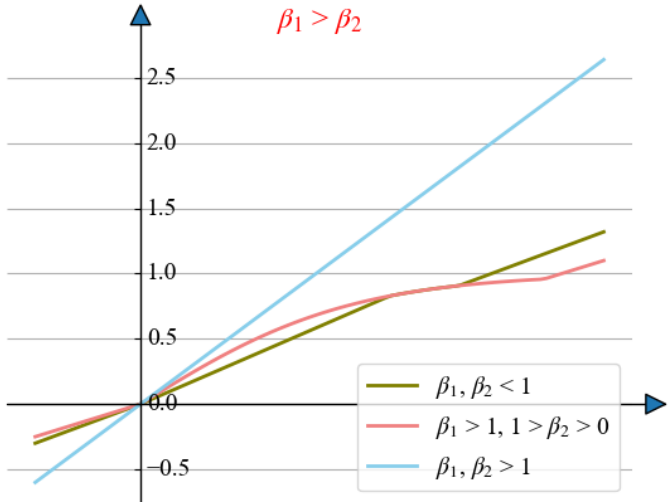

Fig.11: The trend of KDAC when $\beta_1 > \beta_2$.

Next, we analyze the parameter $\mu$. $\mu$ can control the range of non-differentiable points that need to be smoothed. The following are the details.

$\mu$ is established by calculating the intersection of two functions, which determines the range of intersection to be smoothed. It is easy to see that the larger $\mu$, the more obvious the interference these two functions receive, when $\mu$ approaches the limit 0, they are equivalent to not being processed. Fig.12 is an illustration with ten different specific $\mu$, where, in order to explore the effect of $\mu$ on KDAC, other parameters must be fixed, here we take $\beta_1 = 0.8$ and $\beta_2 = 0.5$. So KDAC has three smooth points, according to the size order of their abscissa, they are generated at the intersection of, $\beta_2 x$ and $\beta_1 x$, $\beta_1 x$ and Tanh, Tanh and $\beta_2 x$ respectively. For clarity, Fig.12 records the second smoothing point.

Obviously, the larger $\mu$ is, the smoother the intersection is, and on the contrary, the sharper it is. On the one hand, it cannot be ignored that the increase of $\mu$ gradually interferes with original functions, when $\mu = 0.5$, see the red curve in Fig.12, $\beta_1 x$ and Tanh have greatly deviated from the original trend, which may weaken the benefits of linearity and exponential, and have negative consequences on the performance of KDAC. On the other hand, the sharper the intersection, the less effective the smoothing operation may be, see the curve when $\mu = 0.0001$.

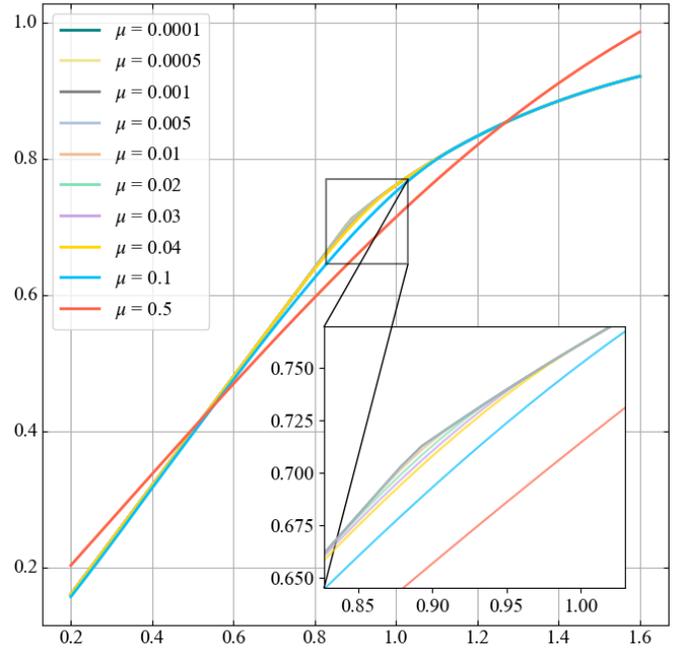

Fig.12: The effect of parameter $\mu$ on KDAC.

To preliminarily verify the above views, we take *HAZOP* dataset (a dataset containing industrial safety knowledge, see Section 4.2 for details) as an example on BERT-BiLSTM-CNN-CRF model (see Section 4.3 for details) for trial experiment and record the F1-score (F1), see Table 1 and Fig.13, where, the "Test" refers to the test set, the "Dev" refers to the validation set.

Table 1: The trial result of $\mu$.

| $\mu$ | F1 (%) | | $\mu$ | F1 (%) | |
| --- | --- | --- | --- | --- | --- |
| | Test | Dev | | Test | Dev |
| 0.0001 | 85.55 | 85.40 | 0.02 | 86.77 | 86.55 |
| 0.0005 | 85.59 | 85.39 | 0.03 | 86.70 | 86.50 |
| 0.001 | 86.28 | 85.94 | 0.04 | 86.55 | 86.39 |
| 0.005 | 86.16 | 86.28 | 0.1 | 86.51 | 86.19 |
| 0.01 | 86.93 | 86.72 | 0.5 | 86.32 | 86.11 |

It can be found that when $\mu = 0.01$, the performance of KDAC reaches the best. When $\mu > 0.01$, with the increase of $\mu$, the performance on both the test set and the verification set shows a continuous decline, and the role of KDAC is gradually weakened. When $\mu < 0.01$, the performance has fluctuated, for example, on the test set, the performance at $\mu = 0.001$ is higher than that at $\mu = 0.005$, but the performance shows a downward trend as a whole.

This confirms our views that the parameter $\mu$ should not be too large, which may affect the diversity of input processing (i.e. different knowledge), nor too small, which may not fully achieve the benefits of smoothing. Therefore, KDAC takes $\mu$ as 0.01 to achieve an appropriate equilibrium.

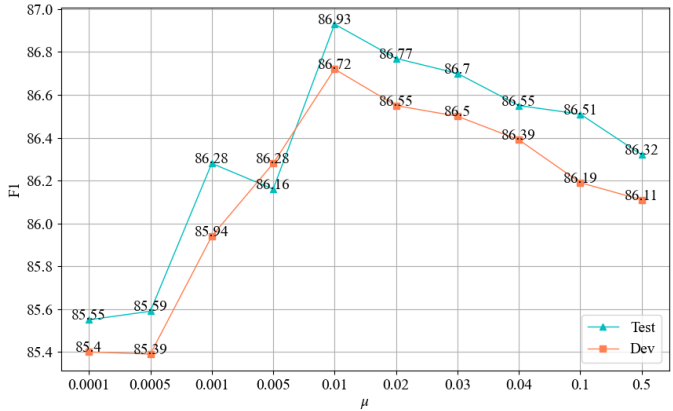

Fig.13: The trial evaluation of $\mu$.

### 3.2.5. Derivative analysis

The first derivative of the activation function is directly related to the back-propagation algorithm, the performance of deep learning is damaged by the lack or saturation, an appropriate activation function should have the ability to surmount these problems. There is no doubt that based on the above analysis, KDAC has these capabilities. To further deepen, we have supplemented some arguments.

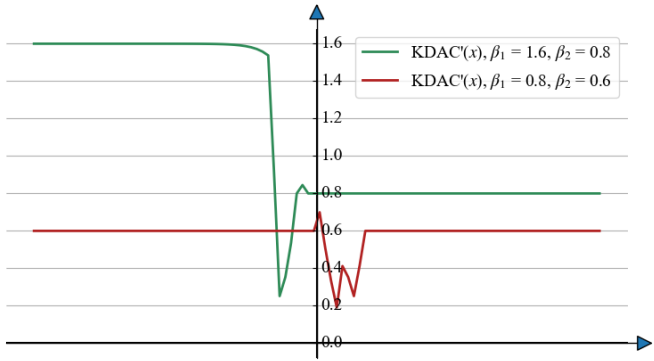

Fig.14: The illustration of two sets of the first derivative of KDAC.

One fact is that the asymptote of Tanh is $y = \pm 1$, so when $x < -1$ or $x > 1$, the absolute value of $\beta_i x$ is greater than that of Tanh, thus, the negative limit and positive limit of KDAC tend to be linear functions, and there is no gradient vanishing problem. Further, the above analysis shows that when $x \rightarrow +\infty$, KDAC $\rightarrow \beta_2 x$, and when $x \rightarrow -\infty$, KDAC $\rightarrow \beta_1 x$ or $\beta_2 x$ (whether $\beta_1 x$ or $\beta_2 x$ depends on the range of their values). Hence, with the increase of input, the first derivative of KDAC is stable as $\beta_1$ or $\beta_2$ in the negative region and $\beta_2$ in the positive region, these two definite constants can ensure that the output value does not change violently with the fluctuation of input value, KDAC can extend anticipately regardless of the size of input argument in various gradients.

Fig.14 is the derivative function diagram of KDAC under the condition of ($\beta_1 = 1.6$, $\beta_2 = 0.8$) and ($\beta_1 = 0.8$, $\beta_2 = 0.6$), there is no fracture in the whole gradient, which is consistent with the above theoretical analysis. So, KDAC has no gradient vanishing and non-differentiable existence, which helps to save the incomplete representation of semantics.

### 3.3. Summary

We have completed the deduction and analysis of KDAC. KDAC benefits from exponential form and linear form, based on the approximate smoothing algorithm, in a clever design way. It has excellent properties, the nonlinearity ensures the fitting ability, the multiple morphological conversion can cope with different inputs, the stable near-linearity transformation and derivative can get rid of gradient vanishing and no negative output, and the approximate smoothing attribute can eliminate non-differentiability.

We outline the structure of the activation function, which is conducive to the overall grasp of the research. What needs attention is that KDAC is not simply a variant, but more the perception and fusion of multiple activation modes. At present, a large number of relevant studies are variants of the existing activation function, and have achieved the expected performance, a common reason is that the variants have some guarantees and interpretability. Specifically, a latest progress has summarized the activation function [31]: ReLU-based variants are still the mainstream, such as APL, FReLU and SReLU; Sigmoid-based variants are also popular, such as SiLU, E-Swish and AGSig; Superimposed variants of Sigmoid and ReLU contain FTS, Swish and Logish [41]; ReLTanh and ABU are based on Tanh and ReLU; AGTanh and Mish are two Tanh-based variants. To be exact, these activation functions mainly follow the linear structure of ReLU or the exponential structure of Tanh and Sigmoid. Furthermore, the activation functions of multilinear-based structures include LiSA [42], RSigELUD [43] and so on. Therefore, KDAC is additionally supported by empirical rationality in structure.

In conclusion, we reiterate that KDAC has the following appreciations and expectations:

[1] Its nonlinearity, near-linearity transformation, multiple form conversion, and derivability can be is expected to respond to different knowledge.

[2] It can conquer the gradient vanishing problem laconically, which is conducive to complete semantic information.

[3] It can relieve the dilemma of no negative output and non-differentiable existence competently, which can assist to compensate for the omission of potential semantic information.

## 4. EXPERIMENT & ANALYSIS

We follow the recent DNER studies to launch experiments, including the selection of evaluation metrics, the type of datasets and the design of models.

### 4.1. Metrics

It is worth noting that the representative studies only take P, R and F1 as performance metrics, even in some complex cases, only F1 is retained for brevity (see the researches mentioned and the investigation in the introduction), because it is sufficient for the evaluation. Where, P is the precision, which represents the proportion that is actually true in the sample predicted to be true, R is the recall, which represents the proportion of samples that are actually true have been excavated, F1 is F1-score, the harmonic mean of P and R, for balancing them, which is more impartial. Formally, we chime in with the above, and report P, R and F1 in this study.

### 4.2. Datasets

In view of the types of datasets involved in Appendix 1, such as social, news and clinical, etc., we perform experiments on six benchmark datasets in distinct domains, including *Weibo* [44], *Clinical* [45], *E-commerce* [46], *People's daily* [47], *Resume* [48] and *HAZOP* [49, 50]. These datasets are composed of different domain knowledge, and the details are as follows.

[1] *Weibo*: It is the category in the social domain, mainly composed of the content various user posts with a total of seven label types, including entity knowledge such as organization, location, and polity, etc.

[2] *Clinical*: The Chinese clinical dataset, constructed by China conference on knowledge graph and semantic computing (CCKS2019), including six entity knowledge types: disease diagnosis, anatomical location, imaging examination, laboratory test, operation and medicine.

[3] *E-commerce*: It belongs to the domain of e-commerce with the types of brands and products.

[4] *People's daily*: It consists of news and has three kinds of entity knowledge: location, organization and person.

[5] *Resume*: It is built from resume information and involves eight types of knowledge entities: education, name, race, country, location, profession, title and organization.

[6] *HAZOP*: The industrial safety dataset from the hazard and operability analysis text, consists of five types of entities: equipment, process label, material, consequence and attribute.

### 4.3. Models

In terms of models, BiLSTM, CRF, CNN and BERT are still great hot (see Section 2.2), different combinations between them are associated with different tasks. Therefore, to be neutral, we combined them without modification, that is, BERT-BiLSTM-CNN-CRF with both the recurrent neural network and the convolutional neural network, as the experimental model, see Fig.15.

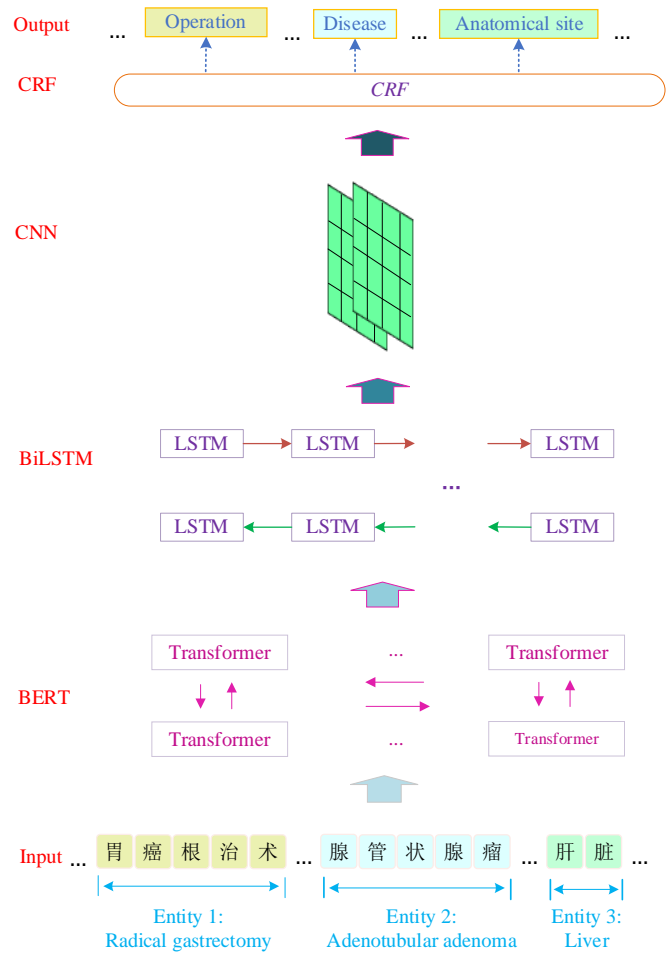

Fig.15: The architecture of the experimental model BERT-BiLSTM-CNN-CRF. Here, we illustrate it with *clinical* data.

We take the average of the results obtained by repeating 5 times for each experimental trial, and we follow the common practice and simply change the activation function in the network. In particular, all other implementation conditions are the same except the activation function in each round of dataset verification. We apply Adam optimizer with the learning rate 1e-3, we use the BERT-Base with 12 hidden layers and 768 hidden size. Besides, we train 20 epochs for each experiment.

## 4.4. Comparisons:

Sigmoid, Tanh and ReLU have led knowledge discovery, so, the three are baselines. The linearity of KDAC is influenced by Leaky ReLU (see Equ.17), and SELU (see Equ.18) also exists in the latest work, thus they are included in the comparison. Besides, Swish [51], LiSA ($\alpha_1 = 0.25$, $\alpha_2 = 0.15$) [42] and RSigELUD ($\alpha = 0.05$, $\beta = 0.2$) [43] are also considered for comparative experiments, although the research they belong to is irrelevant to DNER, they enjoy some similar components to KDAC in structure. Specifically, Swish can be regarded as a representative of activation function smoothing, LiSA is a typical multi-linearity activation function for visual representation, RSigELUD a superposition of ReLU and Sigmoid, see Equ.19, Equ.20 and Equ.21, respectively.

$$Leaky\ ReLU(x) = \begin{cases} x, x > 0 \\ \alpha x, x \leq 0 \end{cases} \qquad (17)$$

$$SELU(x) = \lambda \begin{cases} x, x > 0 \\ \alpha(e^x - 1), x \leq 0 \end{cases} \qquad (18)$$

$$Swish(x) = x \cdot Sigmoid(x) \qquad (19)$$

$$LiSA(x) = \begin{cases} \alpha_1 x - \alpha_1 + 1, x > 1 \\ x, 0 < x < 1 \\ \alpha_2 x, x < 0 \end{cases} \qquad (20)$$

$$RSigELUD(x) = \begin{cases} \alpha \cdot x(1/(1+e^{-x})) + x, x > 1 \\ x, 0 < x < 1 \\ \beta(e^x - 1), x < 0 \end{cases} \qquad (21)$$

## 4.5. Results

Table 2 shows the F1 of the evaluation experiment results, where, in each dataset, "Test" refers to its test set, "Dev" refers to its validation set, and the following numbers report F1 for the total entity. In addition, for more intuitive analysis and discussion, we have prepared a series of figures. There are the following major observations.

Macroscopically, KDAC has taken the lead in 8 of the 12 performance evaluations (see Table 1). It is worth noting that KDAC completely occupies *HAZOP* and *People's daily*, more surprisingly, it does not discard other datasets, since for each set of test sets and verification sets, it obtains the optimum in one of them. KDAC is superior to other activation functions in generalization, which can treat various knowledge and serve the application of knowledge in different fields with a more effective attitude.

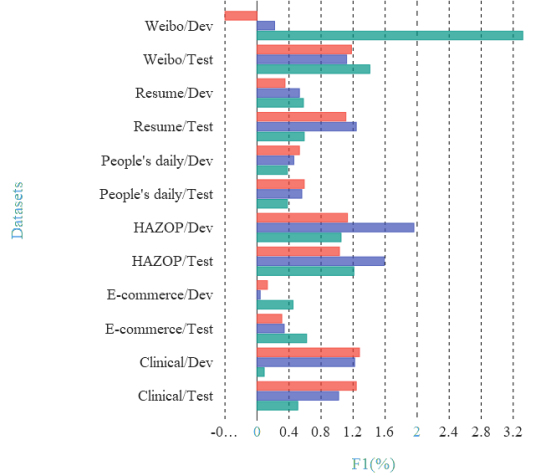

Fig.16: Compared with Tanh, Sigmoid and ReLU that dominate DNER, the increase of F1 under KDAC, where, take "KDAC / ReLU" as an example, which refers to the performance gain of KDAC compared with ReLU. Likewise, this is also consistent with the following figures.

Compared with Tanh, Sigmoid and ReLU that dominate DNER, F1 of DNER under KDAC activation is higher than that under their activation on all datasets except the validation set in *Weibo* under Sigmoid (see Fig.16). Further, KDAC greatly outperforms ReLU on *Weibo* and *HAZOP* and slightly superior to ReLU on other datasets. Specifically, for F1 on *Weibo*, KDAC exceeds ReLU 1.41% and 3.32% in test set and verification set respectively, and has 1.21% and 1.05% improvement on *HAZOP* respectively. This may be the omission of latent semantics induced by no negative output, and is more intense in the domain knowledge that is relatively hard to understand by DNER, such as *Weibo* and *HAZOP*, their F1 fluctuates around 50% and 85% respectively, see Fig.17 and Fig.18. Besides, KDAC also leads Sigmoid and Tanh in performance, it is at least 1.5% higher than Tanh on *HAZOP* and is about 1.2% higher than Sigmoid on *Clinical*, and has a slight lead on almost all other datasets, which can explain that KDAC indirectly improves the complete representation of semantics by overcoming the obstacle of gradient vanishing in a feasible way. These brilliant manifestations encourage KDAC to break through the prison composed of the three activation functions and be promoted a more effective activation function for knowledge discovery.

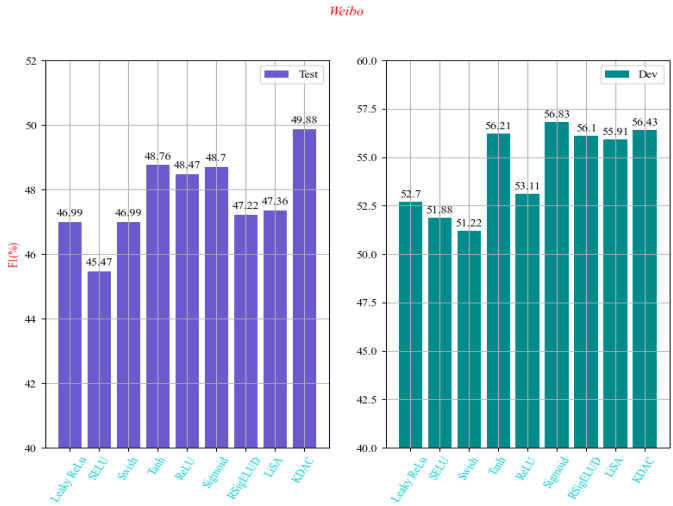

Fig.17: The performance of activation functions on *Weibo*.

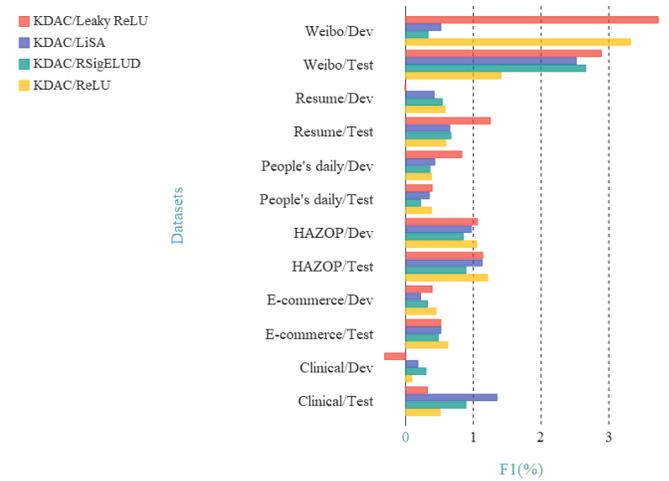

Fig.19: The excitation under KDAC about eliminating non-differentiable points.

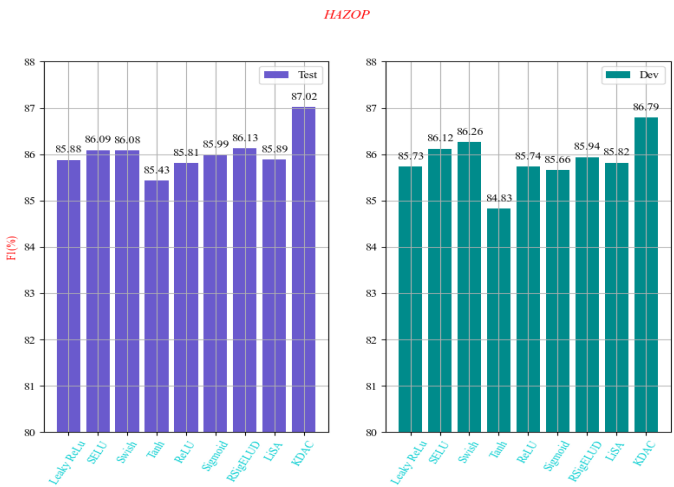

Fig.18: The performance of activation functions on *HAZOP*.

Fig.19 exhibits the incentive under KDAC about eliminating non-differentiable existence, where, Leaky ReLU, LiSA and RSigELUD have no the neuronal saturation, but enjoy non-differentiable points due to function segmentation. F1 of DNER under KDAC activation is higher than that under the three activation on all datasets except the validation set in *Clinical* and *Resume* under Leaky ReLU (see Fig.20 and Fig.21 for details), which indicates that KDAC is effective and advanced. Furthermore, it can be found that the gain amplitude of KDAC on *Weibo* is relatively large, and that on *HAZOP* cannot be underestimated, which may mean that the restriction of non-differentiable points may increase with the rise of the difficulty in understanding knowledge. We detect the non-differentiable points and enable KDAC to consistently and reasonably eliminate it, which is credible and reliable in knowledge exploration.

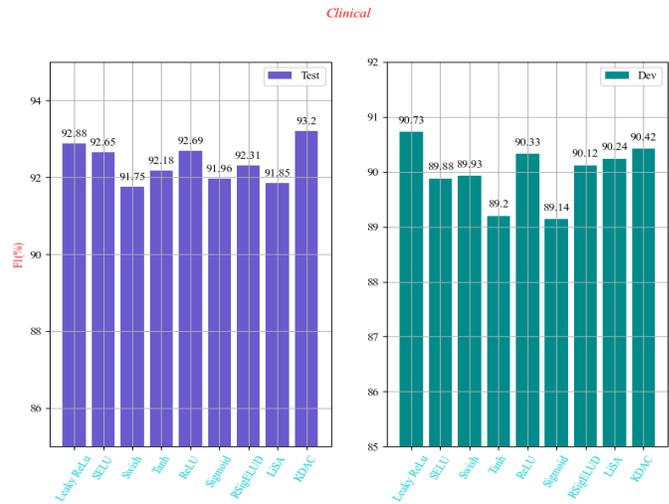

Fig.20: The performance of activation functions on *Clinical*.

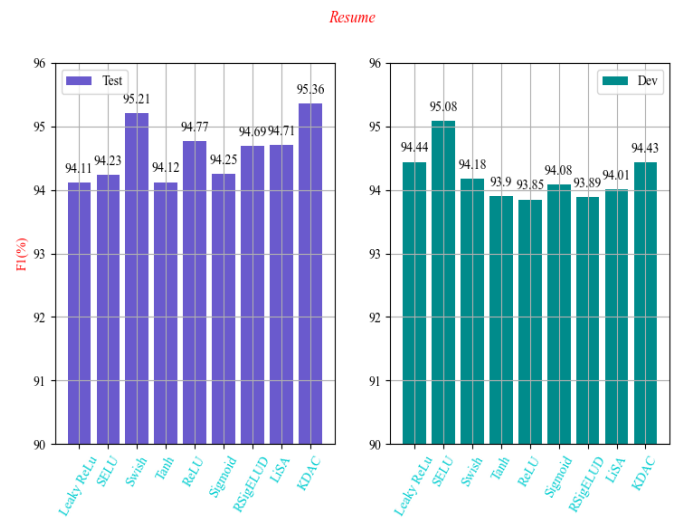

Fig.21: The performance of activation functions on *Resume*.

In terms of smoothing, KDAC improves the performance of DNER remarkably compared with Swish on almost all domains (see Fig.22) except the test set in *E-commerce* (see Fig.23 for details). Likewise, for *Weibo*, the dataset that is difficult to radiate by other activation functions, Swish is no exception, KDAC is a lot of percentage points ahead relatively on both test set and verification set. One possible reason is that the gradient of Swish disappears in the negative region, even if it is at liberty in the positive region, which can indirectly reflect that KDAC is beneficial in surmounting the gradient vanishing. Additionally, Table 1 also shows that the performance of Swish in activating DNER is close to that of Tanh and Sigmoid, which may be affected by the threshold of gradient vanishing. We speculate that this is one of the reasons why previous studies did not leverage Swish as the activation substitute, and may also be the equivalence that SELU failed to become the mainstream, after all, the primary goal of knowledge discovery that usually doesn't care about timeliness is to improve the correctness of entity recognition.

For conciseness, we report the P and R of the experimental results in Appendix 2 and Appendix 3, respectively. Fig.24 illustrates the performance gain in P brought by KDAC to DNER compared with other activation functions, and Fig. 25 is the performance gain comparison in R. The results show that, with a few exceptions, the performance of KDAC is more robust than other activation functions, obviously, this is consistent with the results on F1 (not be repeated here), which also indicates that KDAC has brilliant generalization, and is advanced and reliable.

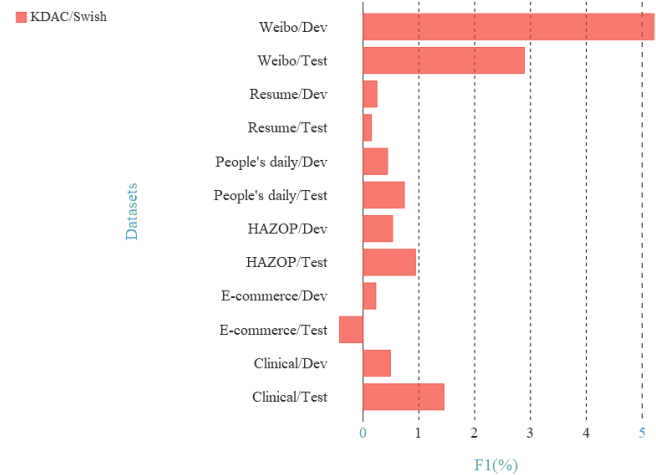

Fig.22: KDAC and Swish about smoothing.

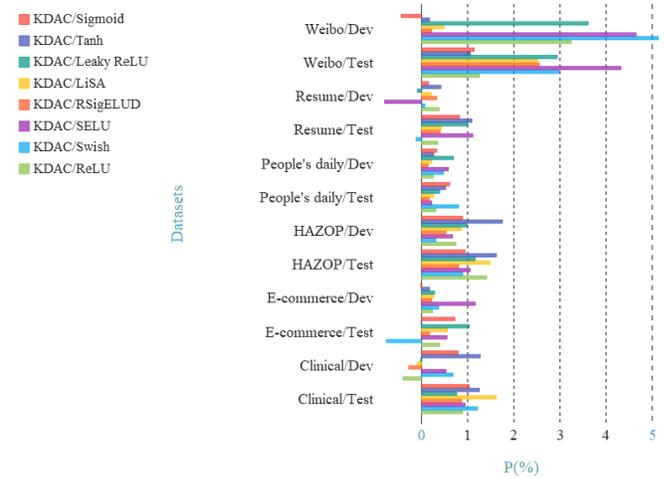

Fig.24: The performance gain in P brought by KDAC to DNER compared with other activation functions.

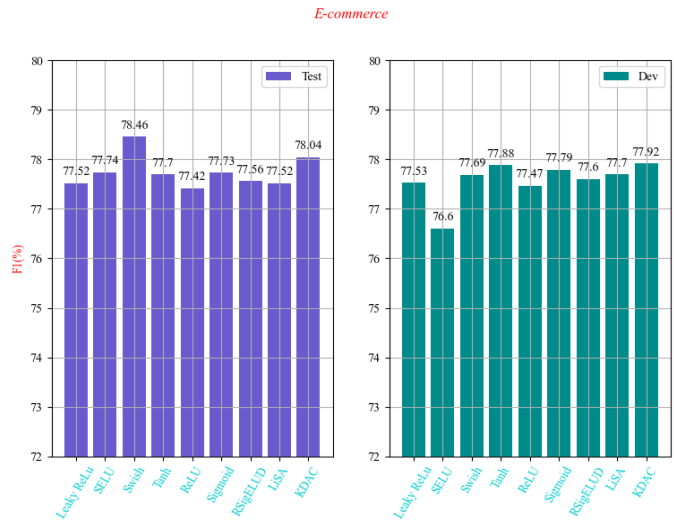

Fig.23: The performance of activation functions on *E-commerce*.

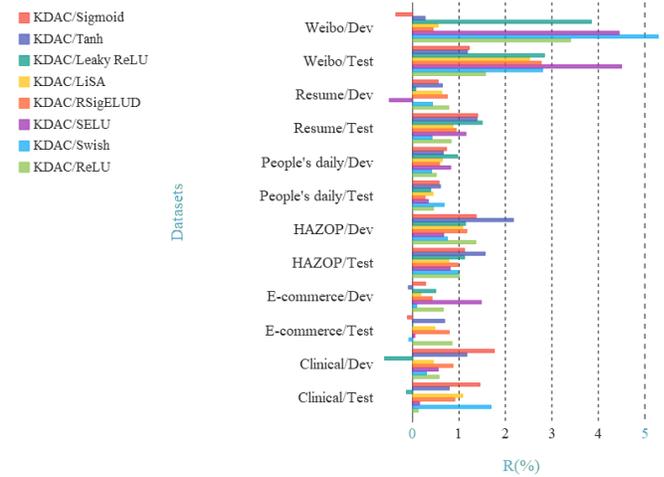

Fig.25: The performance gain in R brought by KDAC to DNER compared with other activation functions.

Table 2: Results of evaluation experiment in F1.

| Dataset<br>Activation | Clinical | | E-commerce | | HAZOP | | People's daily | | Resume | | Weibo | |
|---|---|---|---|---|---|---|---|---|---|---|---|---|
| | Test | Dev | Test | Dev | Test | Dev | Test | Dev | Test | Dev | Test | Dev |
| Leaky ReLU | 92.88 | **90.73** | 77.52 | 77.53 | 85.88 | 85.73 | 93.82 | 94.49 | 94.11 | 94.44 | 46.99 | 52.70 |
| SELU | 92.65 | 89.88 | 77.74 | 76.60 | 86.09 | 86.12 | 93.93 | 94.62 | 94.23 | **95.08** | 45.47 | 51.88 |
| Swish | 91.75 | 89.93 | **78.46** | 77.69 | 86.08 | 86.26 | 93.47 | 94.88 | 95.21 | 94.18 | 46.99 | 51.22 |
| Tanh | 92.18 | 89.20 | 77.70 | 77.88 | 85.43 | 84.83 | 93.65 | 94.86 | 94.12 | 93.90 | 48.76 | 56.21 |
| ReLU | 92.69 | 90.33 | 77.42 | 77.47 | 85.81 | 85.74 | 93.83 | 94.94 | 94.77 | 93.85 | 48.47 | 53.11 |
| Sigmoid | 91.96 | 89.14 | 77.73 | 77.79 | 85.99 | 85.66 | 93.62 | 94.79 | 94.25 | 94.08 | 48.70 | **56.83** |
| RSigELUD | 92.31 | 90.12 | 77.56 | 77.60 | 86.13 | 85.94 | 93.99 | 94.96 | 94.69 | 93.89 | 47.22 | 56.10 |
| LiSA | 91.85 | 90.24 | 77.52 | 77.70 | 85.89 | 85.82 | 93.86 | 94.89 | 94.71 | 94.01 | 47.36 | 55.91 |
| KDAC | **93.20** | 90.42 | 78.04 | **77.92** | **87.02** | **86.79** | **94.21** | **95.32** | **95.36** | 94.43 | **49.88** | 56.43 |

## 5. DISCUSSION

KDAC is exquisite, it skillfully breaks through the dilemma faced by the previous activation functions, steadily promotes the performance of DNER and can serve knowledge discovery with a more promising attitude.

One petty thing is that the multiple form transformation of KDAC makes it have relatively large time complexity. Specifically, the mathematical expression of KDAC is the nesting of $P_{Min}$ and $P_{Max}$, both of which are quadratic polynomials, and their time complexity is $O(n^2)$. Further, the time complexity of Tanh can be regarded as $O(e^{2n})$. Hence, the time complexity of KDAC is $O(e^{8n})$, which is a weakness that more time is consumed when information is mapped through KDAC. Fortunately, the knowledge discovery in DNER is commonly offline and insensitive to timeliness, it mainly pursues the knowledge recognition performance as high as possible, since the importance of the correctness is higher than that of the speed.

From another point of view, the " No Free Lunch Theorem " in AI can indicate that this is an extremely common and reasonable result [52], because nothing is perfect. While pursuing high performance, algorithms or models frequently involuntarily slow down their speed, and vice versa, they sacrifice some performance to improve their speed in some high real-time projects. For the former, such as BERT family and GPT family, they promote the prediction task and generation task of NLP respectively with excessive parameters and complexity. The latter includes knowledge distillation algorithm.

All in all, KDAC with sufficient performance is competent for knowledge discovery.

## 6. CONCLUSION

DNER is the paradigm of knowledge discovery, which can promote the prosperity of the current data era. However, the activation functions embedded in DNER sink into the dilemma of gradient vanishing, non-differentiability or no negative output, which may cause the omission of features and the incomplete representation of latent semantics.

To surmount these non-negligible barriers, we propose a novel activation function termed KDAC. KDAC draws inspirations from linear Newton interpolation for nonlinear eigenvalue problem, and leverages an approximate smoothing algorithm to manage non-differentiable existence. Besides, KDAC absorbs Tanh form exponential structure and adjustable linear structure to treat gradient vanishing and no negative output. KDAC has a series of gratifying properties. It can activate various modes, change dynamically and freely convert diverse inputs, to respond to different types of knowledge, its nonlinearity ensures the fitting ability, its near-linear transformation and derivability can guarantee the stability of knowledge mapping.

To fairly evaluate KDAC, we have selected six datasets with different domain knowledge, such as *Weibo*, *Clinical*, *E-commerce*, *Resume*, *HAZOP* and *People's daily*. In addition, we have employed the prevalent activation functions including SELU, Swish, Tanh, ReLU, Sigmoid, RSigELUD, Leaky ReLU and LiSA for comparison. The evaluation model is BERT-BiLSTM-CNN-CRF. The experimental results show that KDAC is superior, advanced and efficient, and can provide more generalized activation to stimulate the performance of DNER. KDAC can be exploited as a promising alternative activation function in DNER to devote itself to the knowledge discovery. Luckily, the high time complexity of KDAC weakly affects DNER. Additionally, to the best of our knowledge, there are no relevant works dedicated to this research, and our work comes to fill in this gap.

In the future work, we will study how to boost DNER from other aspects. We expect that our work can promote the prosperity of knowledge discovery and its application.

## ACKNOWLEDGEMENTS

This work was supported by the National Natural Science Foundation (NNSF) of China (61703026). Additionally, we are particularly grateful for the insightful suggestions from Dong Gao, as well as the technical support from Fanglin Liu and Haozhe Liu.

## DECLARATION OF COMPETING INTEREST

The authors declare that they have no known competing financial interests or personal relationships that could have appeared to influence the work reported in this paper.

# APPENDIX

Appendix 1: The survey of the latest representative DNER.

| Journal | Activation functions | Main models | Main contributions | Main datasets | Reference |
|---|---|---|---|---|---|
| *Neu* | Tanh, ReLU | CNN, BiLSTM, CRF, VGG-16 | Multi-modal based | Twitter | 53 |
| | / | BiLSTM | Nested entities based | ACE2004, ACE2005, GENIA | 54 |
| | ReLU | CNN, BiLSTM | Onion domains of the Tor Darknet | WNUT2017 | 55 |
| | Tanh | BiLSTM, CRF | Token-level based | CoNLL-2003 | 56 |
| | / | BERT, CRF | Chemicals management | Chemical safety | 57 |
| | Tanh | BiLSTM, CRF | Nested entities based | ACE2005, CoNLL2003, GENIA | 58 |
| | Tanh | BiLSTM, CRF | The identification of adverse drug reactions | Drug | 59 |
| | ReLU | ODE-net | The rumor detection | Twitter, Weibo | 60 |
| | Tanh | CNN, BiLSTM, CRF | The medical diagnosis | i2b2/VA, E-medical record | 61 |
| | Tanh | CNN based, BiLSTM | Joint entity and relationship extractions | NYT10, NYT11 | 62 |
| | Tanh, Sigmoid | LSTM, Transformer | Joint entity and relationship extractions | NYT24, NYT29 | 63 |
| | SELU | CNN, Attention | Joint entity and relationship extractions | NYT, WebNLG | 64 |
| | Tanh | LSTM, CRF | CRF variants | Conll2003, Conll2000, BC2GM, JNLPBA, CHEMDNER | 65 |
| | Tanh, ReLU | BiLSTM, CRF | Nested entities based | ACE2005 | 66 |
| *NN* | Sigmoid | BiLSTM, CRF | Nested entities based | GENIA, JNLPBA | 67 |
| | / | CNN, BiLSTM, CRF, Attention | Cross-domain based | CoNLL-2003、Twitter、SciTech | 68 |
|  | / | CNN, BiLSTM, CRF | Transferable based | OntoNotes 5.0, Ritter11, GUM, MIT movies | 69 |
|  | Sigmoid | BiLSTM, BiGRU | The medical diagnosis | I2B2 2009 | 70 |
| ESA | Tanh | BiLSTM | The Indonesian | Indonesian | 71 |
|  | Sigmoid | / | The geographic exploration | Mexican news | 72 |
|  | / | BiLSTM, Transformer, CRF | The Turkish | Milliyet | 73 |
|  | ReLU | CNN, BiLSTM, CRF, VGG-16 | Multi-modal based | CoNLL 2003, Twitter | 74 |
|  | Sigmoid | BERT based | Joint entity and relationship extractions | NYT, WebNLG | 75 |
| IPM | ReLU | CNN, BiLSTM, CRF | Diabetes care | Chinese Wikipedia, Diabetes | 76 |
|  | / | Transformer | The gender bias detection | CoNLL-g, Wiki-g, IEER-g, Textbook-g | 77 |
|  | Tanh | BERT, BiLSTM | The event extraction | ACE2005 | 78 |
|  | / | BERT, BiLSTM, CRF | Nested entities based | Turkish bank documents | 79 |
| KBS | ReLU | CNN, BiGRU | Token-level based | IJCNLP-08, NERSSEAL, Punjabi language | 80 |
|  | Sigmoid | BERT, BiLSTM, CRF | Car reviews | Car reviews | 81 |
|  | Tanh | BERT, BiLSTM, CRF, Attention | The biochemical | CHEMDNER | 82 |
|  | / | CNN_gated, BiLSTM, CRF | Adversarial trained | CoNLL-03, OntoNotes 5.0, WNUT-17 | 83 |
|  | / | CNN, BiLSTM, CRF | Scientific literatures | Scientific literatures | 84 |
|  | / | BiLSTM, CRF | The clinical identification | i2b2/UTHealth 2014 | 85 |

Appendix 2: Results of evaluation experiment in P.

| Dataset<br>Activation | Clinical | | E-commerce | | HAZOP | | People's daily | | Resume | | Weibo | |
|---|---|---|---|---|---|---|---|---|---|---|---|---|
| | Test | Dev | Test | Dev | Test | Dev | Test | Dev | Test | Dev | Test | Dev |
| Leaky ReLU | 92.33 | 90.47 | 76.88 | 77.62 | 85.93 | 85.66 | 93.79 | 94.52 | 94.17 | 94.39 | 46.91 | 52.83 |
| SELU | 92.15 | 89.92 | 77.36 | 76.74 | 86.04 | 85.97 | 93.96 | 94.63 | 94.06 | **95.10** | 45.53 | 51.79 |
| Swish | 91.88 | 89.77 | **78.67** | 77.53 | 86.20 | 86.33 | 93.38 | 94.74 | **95.29** | 94.23 | 46.87 | 51.31 |
| Tanh | 91.84 | 89.18 | 77.92 | 77.73 | 85.47 | 84.89 | 93.66 | 94.95 | 94.08 | 93.88 | 48.79 | 56.27 |
| ReLU | 92.20 | **90.85** | 77.52 | 77.66 | 85.68 | 85.90 | 93.87 | 94.96 | 94.82 | 93.92 | 48.60 | 53.20 |
| Sigmoid | 92.06 | 89.66 | 77.19 | 77.92 | 86.15 | 85.75 | 93.57 | 94.88 | 94.35 | 94.15 | 48.71 | **56.88** |
| RSigELUD | 92.22 | 90.73 | 77.74 | 77.68 | 86.29 | 86.11 | 94.01 | 95.07 | 94.77 | 93.97 | 47.30 | 56.22 |
| LiSA | 91.47 | 90.54 | 77.35 | 77.64 | 85.61 | 85.78 | 93.92 | 94.99 | 94.74 | 94.09 | 47.33 | 55.95 |
| KDAC | **93.09** | 90.45 | 77.91 | **77.90** | **87.09** | **86.64** | **94.18** | **95.21** | 95.17 | 94.30 | **49.85** | 56.44 |

Appendix 3: Results of evaluation experiment in R.

| Dataset<br>Activation | Clinical | | E-commerce | | HAZOP | | People's daily | | Resume | | Weibo | |
|---|---|---|---|---|---|---|---|---|---|---|---|---|
| | Test | Dev | Test | Dev | Test | Dev | Test | Dev | Test | Dev | Test | Dev |
| Leaky ReLU | **93.44** | **90.99** | 78.17 | 77.44 | 85.83 | 85.80 | 93.85 | 94.46 | 94.05 | 94.49 | 47.07 | 52.57 |
| SELU | 93.16 | 89.84 | 78.12 | 76.46 | 86.14 | 86.27 | 93.90 | 94.61 | 94.40 | **95.06** | 45.41 | 51.97 |
| Swish | 91.62 | 90.09 | 78.25 | 77.85 | 85.96 | 86.19 | 93.56 | 95.02 | 95.13 | 94.13 | 47.11 | 51.13 |
| Tanh | 92.52 | 89.22 | 77.48 | 78.03 | 85.39 | 84.77 | 93.64 | 94.77 | 94.16 | 93.92 | 48.73 | 56.15 |
| ReLU | 93.19 | 89.82 | 77.32 | 77.28 | 85.94 | 85.58 | 93.79 | 94.92 | 94.72 | 93.78 | 48.34 | 53.02 |
| Sigmoid | 91.86 | 88.63 | **78.28** | 77.66 | 85.83 | 85.57 | 93.67 | 94.70 | 94.15 | 94.01 | 48.69 | **56.78** |
| RSigELUD | 92.40 | 89.52 | 77.38 | 77.52 | 85.97 | 85.77 | 93.97 | 94.85 | 94.61 | 93.81 | 47.14 | 55.98 |
| LiSA | 92.23 | 89.94 | 77.69 | 77.76 | 86.17 | 85.86 | 93.80 | 94.79 | 94.68 | 93.93 | 47.39 | 55.87 |
| KDAC | 93.31 | 90.39 | 78.17 | **77.94** | **86.95** | **86.94** | **94.24** | **95.43** | **95.55** | 94.56 | **49.91** | 56.42 |